\documentclass[letterpaper, 10 pt, journal, twoside]{IEEEtran}

\IEEEoverridecommandlockouts                              

\usepackage{amssymb}
\usepackage{amsmath}
\usepackage{graphicx}
\usepackage[table,xcdraw]{xcolor}
\usepackage{booktabs}
\usepackage{multirow}
\usepackage[super]{nth}

\newcommand{\norm}[1]{\left\lVert#1\right\rVert}

\newcommand{\mat}[1]{\mathbf{#1}}

\DeclareMathOperator{\argmin}{\text{arg\,min}}
\newcommand{\tf}[2]{\ensuremath{^{\text{#1}}\mat{T}^{}_{\text{#2}}}}
\newcommand{\rf}[2]{\ensuremath{^{\text{#1}}{R}^{}_{\text{#2}}}}
\newcommand{\pf}[2]{\ensuremath{^{\text{#1}}\mat{t}^{}_{\text{#2}}}}

\title{Reflective-AR Display: An Interaction Methodology for Virtual-to-Real Alignment in Medical Robotics
}

\begin{document}

\author{Javad Fotouhi$^{1,2}$ \and
Tianyu Song$^{2}$ \and Arian Mehrfard$^{2}$ \and Giacomo Taylor$^{1,2,4}$ \and Qiaochu Wang$^{2}$ \and Fengfan Xian$^{2}$ \and Alejandro Martin-Gomez$^{3}$ \and Bernhard Fuerst$^{4}$ \and Mehran Armand$^{5}$ \and Mathias Unberath$^{1,2}$ \and Nassir Navab$^{1,2,3}$
\thanks{Manuscript received: 09, 10, 2019; Revised 12, 21, 2019; Accepted 01, 17, 2020.}
\thanks{This paper was recommended for publication by Editor Pietro Valdastri upon evaluation of the Associate Editor and Reviewers' comments.}
\thanks{$^{1}$ J. Fotouhi, G. Taylor, M. Unberath, and N. Navab are with the Department of Computer Science, Johns Hopkins University, $3400$ N. Charles Street, Baltimore, USA 
        {\tt\footnotesize javad.fotouhi@jhu.edu, gzt@jhu.edu, unberath@jhu.edu, nassir.navab@jhu.edu}}%
\thanks{$^{2}$ J. Fotouhi, T. Song, A. Mehrfard, Q. Wang, F. Xian, G. Taylor, M. Unberath, and N. Navab are with the Laboratory for Computer Aided Medical Procedures, Johns Hopkins University, $3400$ N. Charles Street, Baltimore, USA
        {\tt\footnotesize javad.fotouhi@jhu.edu, tsong11@jhu.edu, amehrfa1@jhu.edu, qwang85@jhu.edu, fxian1@jhu.edu, gzt@jhu.edu, unberath@jhu.edu, nassir.navab@jhu.edu}}%
\thanks{$^{3}$ A. Martin-Gomez and N. Navab are with the Laboratory for Computer Aided Medical Procedures, Technical University of Munich,  $3$ Boltzmannstr, Munich, Germany
        {\tt\footnotesize alejandro.martin@tum.de, nassir.navab@jhu.edu}}%
\thanks{$^{4}$ G. Taylor and B. Fuerst are with Verb Surgical Inc., $2450$ Bayshore Pkwy, Mountain View, USA
        {\tt\footnotesize gzt@jhu.edu, benfuerst@verbsurgical.com}}%
\thanks{$^{5}$ M. Armand is with the Applied Physics Laboratory, Johns Hopkins University,  $11100$ Johns Hopkins Road, Laurel, Maryland, USA
        {\tt\footnotesize mehran.armand@jhuapl.edu}}%
\thanks{Digital Object Identifier (DOI): see top of this page.}
}

\markboth{IEEE Robotics and Automation Letters. Preprint Version. Accepted January, 2020} %
{Fotouhi \MakeLowercase{\textit{et al.}}: Reflective-AR} 

\maketitle

\begin{abstract}

Robot-assisted minimally invasive surgery has shown to improve patient outcomes, as well as reduce complications and recovery time for several clinical applications. While increasingly configurable robotic arms can maximize reach and avoid collisions in cluttered environments, positioning them appropriately during surgery is complicated because safety regulations prevent automatic driving. We propose a head-mounted display (HMD) based augmented reality (AR) system designed to guide optimal surgical arm set up. The staff equipped with HMD aligns the robot with its planned virtual counterpart. In this user-centric setting, the main challenge is the perspective ambiguities hindering such collaborative robotic solution. To overcome this challenge, we introduce a novel registration concept for intuitive alignment of AR content to its physical counterpart by providing a multi-view AR experience via reflective-AR displays that simultaneously show the augmentations from multiple viewpoints. Using this system, users can visualize different perspectives while actively adjusting the pose to determine the registration transformation that most closely superimposes the virtual onto the real. The experimental results demonstrate improvement in the interactive alignment of a virtual and real robot when using a reflective-AR display. We also present measurements from configuring a robotic manipulator in a simulated trocar placement surgery using the AR guidance methodology.

\end{abstract}


\section{INTRODUCTION}
\IEEEPARstart{R}{obotic-assisted} minimally invasive surgery is becoming increasingly common due to its associated benefits that include higher accuracy, and tremor and fatigue reduction. Robotic systems can augment the surgeon's abilities with stereo endoscopic imaging and intuitive control which help the surgeon's hand-eye coordination and reduce physical workload during surgery~\cite{van2009ergonomics}. Furthermore, robotic surgery has benefits over traditional laparoscopic techniques with patients experiencing reduced blood loss and shorter post-operative hospital stays~\cite{jayakumaran2017robotic}. 

\begin{figure}[h]
  \centering
  \includegraphics[width=\columnwidth]{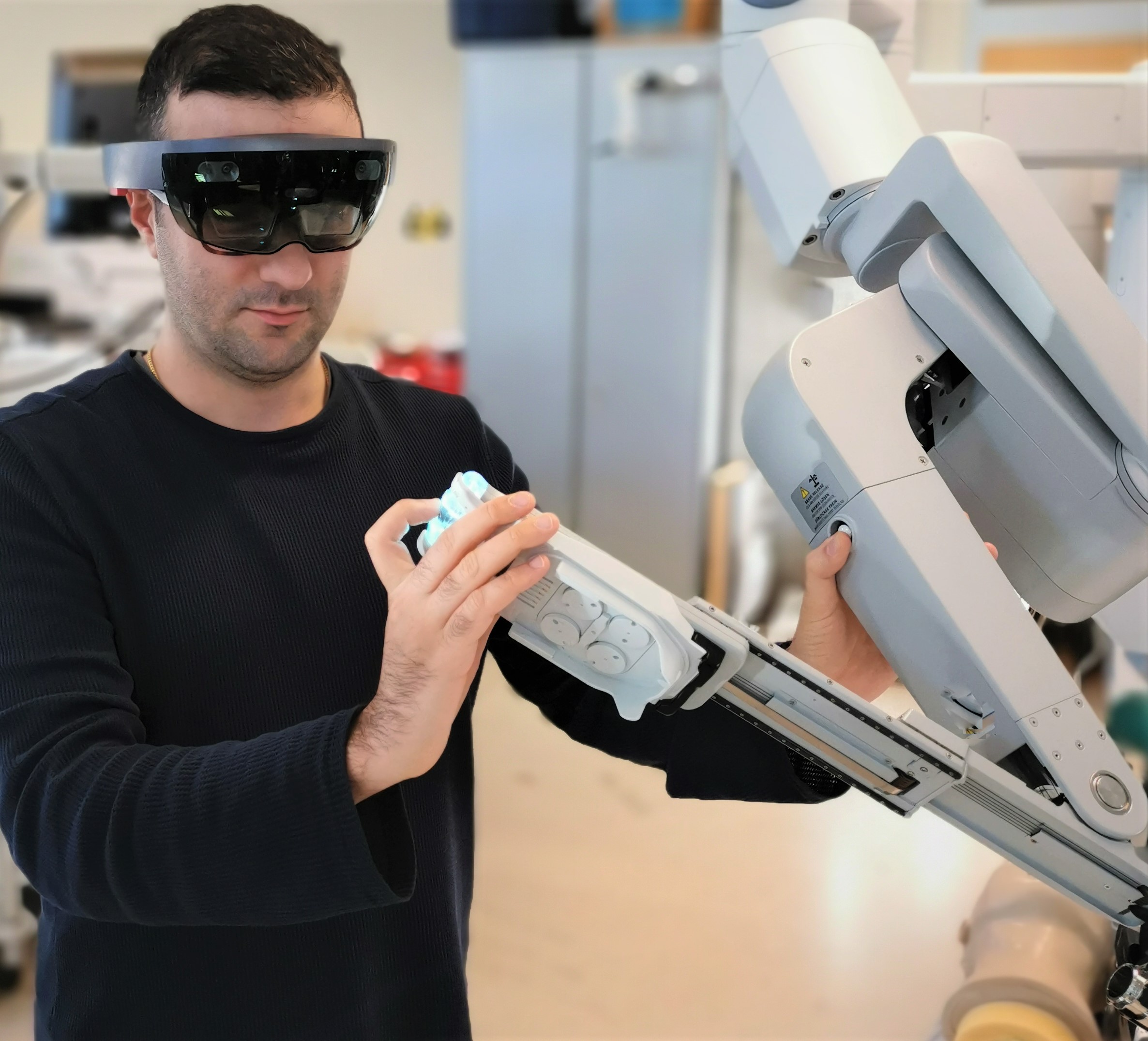}
  \caption{AR-assisted robot arm positioning}
    \label{fig:0}
\end{figure}

Quick and accurate set up of robotic systems leading up to surgery remains a major challenge in the endeavor of making robotic surgery the standard of care. After a patient has been positioned, anesthetized, and trocars inserted, the robotic arms must be positioned and docked before operation can begin. This procedure is a crucial step of workflow and grows more complex the more joints the robotic arms have. While many different configurations of the robot's joints may allow the robotic arm to dock, sub-optimal positioning increases the likelihood of collisions and inadequate reach during teleoperation. Any repositioning or undocking necessary to adjust the robot significantly decreases operating room efficiency~\cite{juza2018intraoperative}. For many procedures including minimally invasive gastrectomy, "junk time", the time taken to set up or reposition robotic arms, is often the sole reason for increased procedure time in robotic procedures over purely laparoscopic approach~\cite{liu2019reasons}.

\begin{figure}[h]
  \centering
  \includegraphics[width=\columnwidth]{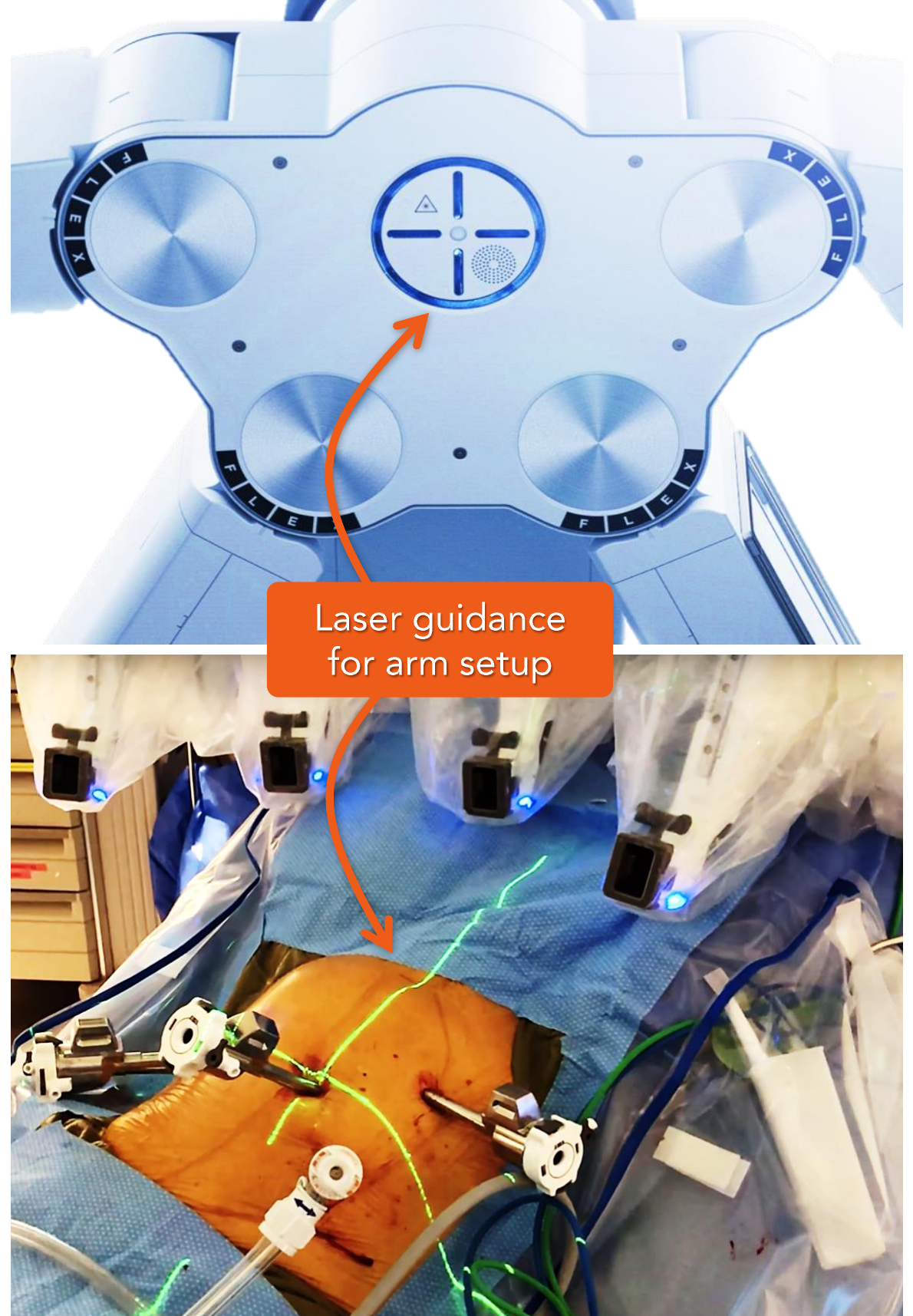}
  \caption{Da Vinci Xi surgical robot uses multiple lasers to assist the positioning and docking of the robotic arms.}
    \label{fig:laser}
\end{figure}

Optimal set up of robotic arms, consequently, is critical to increasing the efficiency of robotic surgery and foster acceptance. Due to safety and regulatory concerns, having a robot automatically drive itself to a pre-operative position is infeasible. In modern surgical robotic systems, the set up of the arms can be supported by lasers, as shown in Fig.~\ref{fig:laser}. Though lasers assist the staff in aligning the robot, it can still exhibit challenges in a complex system with joint redundancies, as it does not directly show the desired configuration of all joints. As most manual methods are error prone and induce a steep learning curve to operating room staff unfamiliar with a system, we investigate an augmented reality (AR) solution for guidance during robotic set up. Using an optical see-through head-mounted display (OST-HMD), setup staff can be interactively guided through joint-by-joint steps to optimally position the robot in an efficient manner.

The works by Qian et al. are similar to our solution in the spirit of using AR for robotic surgery~\cite{qian2019augmented, qian2019aramis, qian2018arssist}; however, the focus of their works were on optimal instrument insertion and manipulation by showing the extension of the arms inside the abdomen using AR. Our methodology addresses the alignment of robot arms for optimal reach and minimum collision. Other early applications of AR in robot-assisted surgery focused on multi-modal registration of medical imaging data with the endoscopic view~\cite{mohareri2013ultrasound, pessaux2015towards}.

Several studies have discussed the challenges of aligning virtual and real objects, and have emphasized the importance of this step for various room-scale and spatially-aware AR solutions~\cite{reiners1998augmented,henderson2011augmented,feiner1993knowledge,caudell1992augmented}. Nuernberger et al. suggested a semi-automatic alignment strategy to register virtual and real spaces~\cite{nuernberger2016snaptoreality}. Their work relied on scene content and environment constraints such as edges and surfaces for snapping the virtual content to real. In a different study, various rendering and visualization techniques were compared for alignment of different virtual models in fully immersive environments~\cite{Martin:2019ab}. Results indicated that static visualization techniques which exhibited lower occlusion in a single view yielded better alignment.

In order to properly augment virtual assistance on a physical robot, we require an intuitive and fast approach to align the AR environment provided by a OST-HMD to the robot. This registration must be robust to perceptual ambiguities that arise during AR alignment~\cite{milgram1997perceptual,willemsen2008effects}. To this end, we propose virtual-real active alignment (ViRAAl) to register a virtual model of the robot to its real counterpart. Our method enables the user to create and view multiple AR mirrors which show the current 3D scene (including real and virtual robot) from different viewpoints. By providing this overlay from multiple perspectives simultaneously, users can actively adjust the 6 degree-of-freedom (DOF) transformation parameters that best align the virtual and real objects in all views. 

\begin{figure}[h]
  \centering
  \includegraphics[width=\columnwidth]{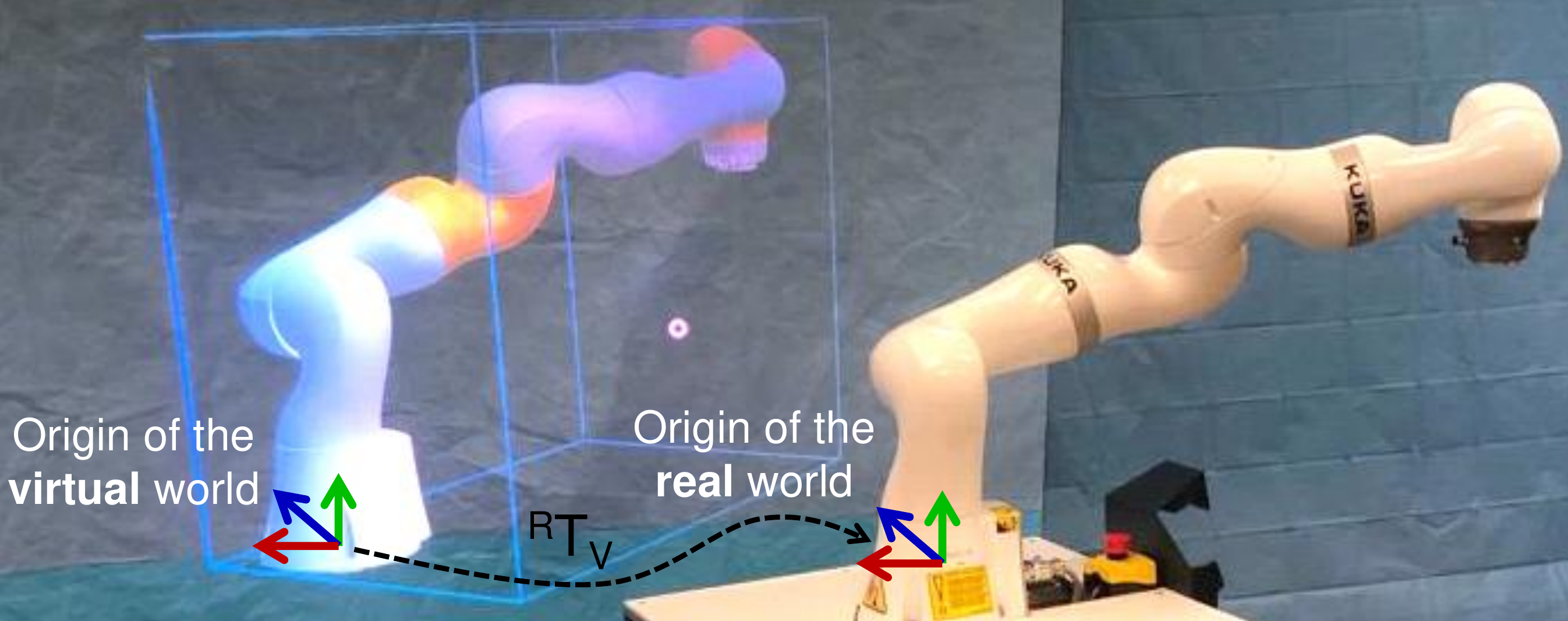}
  \caption{ViRAAl strategy estimates the virtuality to reality transformation $\tf{R}{V}$.}
    \label{fig:1}
\end{figure}

We summarize the contributions of the present work as \textit{1)} reflective-AR displays as the multi-view and marker-free paradigm for co-registration between the virtual and real spaces, hence enabling spatially-aware AR, and \textit{2)} using AR for assistance during robot set up (Fig.~\ref{fig:0}).

\section{METHODOLOGY}
An important step in many AR scenarios is to bring the virtual content that lives in a controlled environment into alignment with the physical reality that is present in the unmodeled environment~\cite{milgram1997perceptual}. In this work, to enable seamless interaction of a surgical robot manipulator and its virtual representation during an AR experience, we introduce reflective-AR displays that enable multi-view visualization and interactive alignment of virtual and real objects. In Sec.~\ref{sec:method:ViRAAl}, we present the problem formulation for registering virtual-to-real. A key contribution of this work, which is the AR reflectors, is presented in Sec.~\ref{sec:method:mirror}. Finally, in Sec.~\ref{sec:method:assistance}, we suggest AR guidance to facilitate robot set up during surgical interventions. It is important to note that in Sec.~\ref{sec:method:ViRAAl} and~\ref{sec:method:mirror} we discuss the problem of \textit{"virtual-to-real"} alignment to enable spatially-aware AR, and in Sec.~\ref{sec:method:assistance} we discuss \textit{"real-to-virtual"} alignment to provide spatially-aware AR guidance.

\subsection{Virtual-Real Active Alignment (ViRAAl)} \label{sec:method:ViRAAl}
To estimate the virtual-to-real $6$ DOF alignment shown in Fig.~\ref{fig:1}, we estimate the transformation $\tf{R}{V} = (\bar{R}, \bar{t})$ via interactively registering a robot with its virtual model at $N$ pre-defined joint configurations. Each time a rigid-body transformation $\left\{ (R_i, t_i) \right\}_{i=1}^{N}$ is obtained, where $(R_i, t_i) \in SE(3)$, and $SE$ is the Special Euclidean group. 

\begin{figure}
  \centering
  \includegraphics[width=\columnwidth]{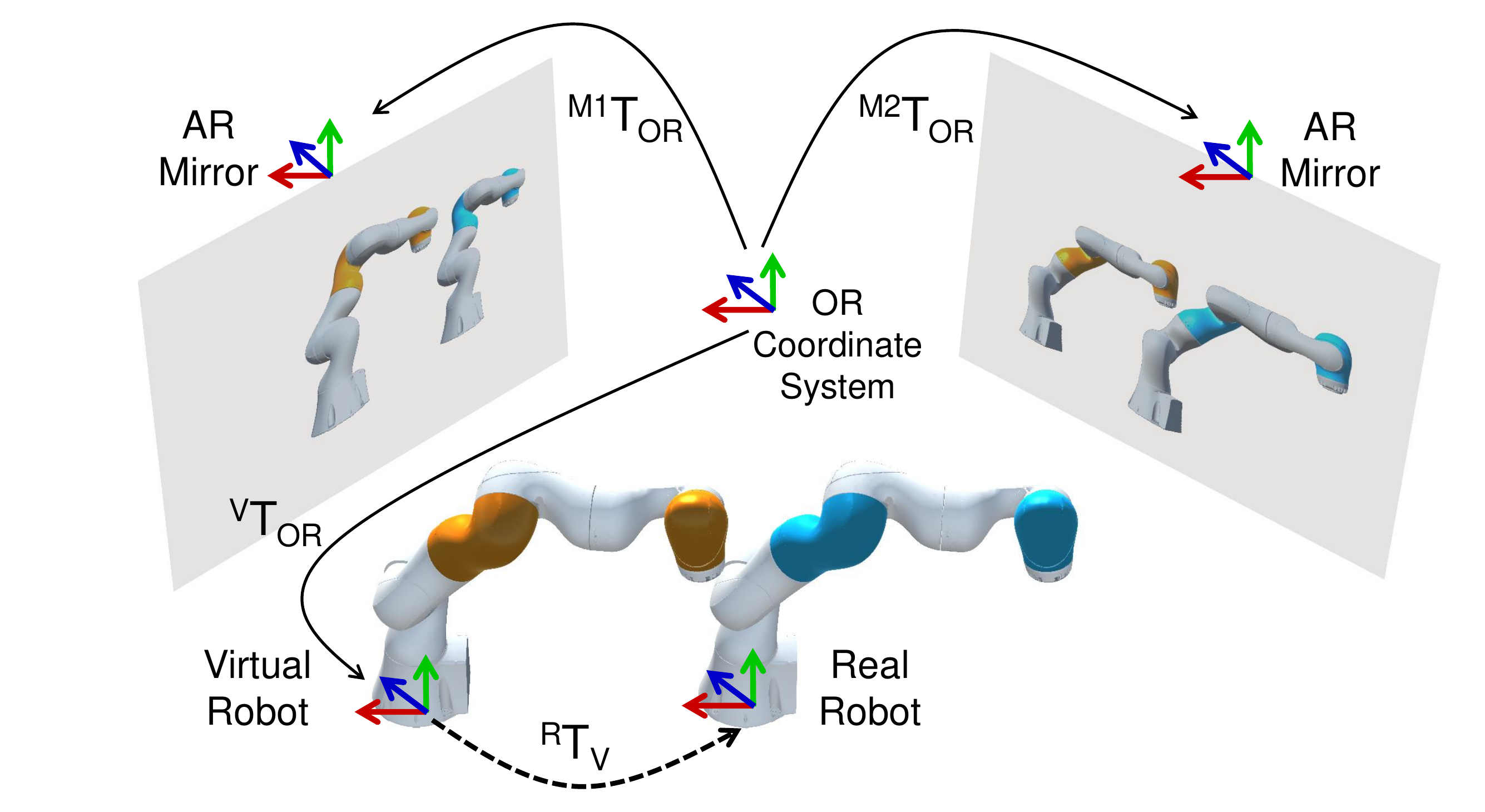}
  \caption{Reflective-AR displays enable simultaneous alignment from multiple views.}
    \label{fig:2}
\end{figure}

\begin{figure*}
  \centering
  \includegraphics[width=0.8\textwidth]{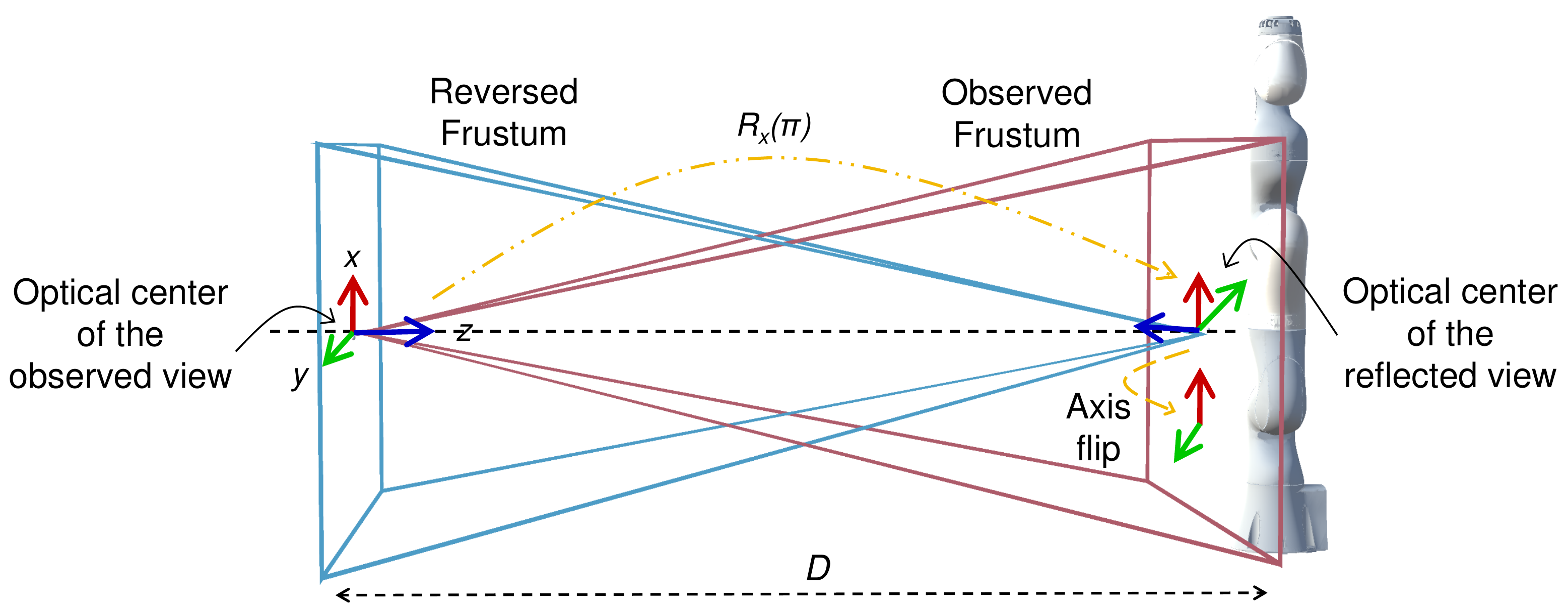}
  \caption{Imaging geometries of the observed and reversed frustums in relation to the robotic manipulator}
    \label{fig:3}
\end{figure*}

We hypothesize that the average transformation computed from these $N$ estimates, can yield a closer approximation of the true virtual-to-real alignment compared to each individual $N$ transformations. Hence, from these $N$ estimates, we seek to compute the mean rotation and translation. The mean rotation matrix $\bar{R}$ is computed on the Special Orthogonal group $SO(3)$ by minimizing:
\begin{equation}\label{eq:averageRotationdefinition}
    \underset{\bar{R} \in SO(3)} \argmin \sum_{i=1}^{N} d(R_i, \bar{R})^2,
\end{equation}
where $d(.)$ denotes a distance function on the Riemannian manifold. To establish $d(.)$, the rotation matrix is expressed in the Lie algebra (tangent space) of the Lie group as $R = e^{\hat{\mat{w}}}$. The tangent space $\mat{w}$ is then obtained as  $\log(R) = \hat{\mat{w}}$, such that $\mat{\hat{w}}$ is the skew-symmetric matrix constructed from the vector $\mat{w}$. Consequently, the mean rotation is estimated as~\cite{moakher2002means}:
\begin{equation}\label{eq:averagerotation}
    \underset{\bar{R} \in SO(3)} \argmin \sum_{i=1}^{N} \norm{\log(R_i^{\top} \bar{R}) }^2_{F},
\end{equation}
where $\norm{.}^2_F$ is the Frobenius norm. The mean translation $\bar{t}$ is computed in Euclidean space as:
\begin{equation}\label{eq:averagetranslation}
    \bar{\mat{t}} = \frac{1}{N} \sum_{i=1}^{N} \mat{t}_i.
\end{equation}

\subsection{Reflective-AR Display} \label{sec:method:mirror}
Due to the projective property of human visual system and the differences in perceptual cues in virtuality and reality, the scale and depth between real and virtual objects are easily misjudged~\cite{willemsen2008effects}. To overcome depth ambiguities and enhance 3D perception during an AR experience, we introduce reflective-AR displays that allow simultaneous visualization of the scene from various viewpoints. The reflective-AR displays shown in Fig.~\ref{fig:2} are constructed by displaying images from the integrated camera sensor of the OST-HMD as if the user observed the real scene from different viewpoints simultaneously, and are augmented with the projections of the 3D virtual objects. 
To compute a geometrically relevant pose for displaying these images, we compute the associated observer poses to the coordinate frame of the AR scene in the operating room $(\rf{O}{OR}, \pf{O}{OR})$ via simultaneous-localization and mapping (SLAM).

The observer imaging geometry in Fig.~\ref{fig:3} is formulated as:
\begin{equation}
    P_{o} = K_o \, P \, \begin{bmatrix} \rf{O}{OR} & \pf{O}{OR} \\ \mat{0}^\top & 1  \end{bmatrix},
\end{equation}
where $K_o$ is the matrix of intrinsic parameters and $P$ is the projection operator. Next, to simulate a mirror-like view, we construct a reversed frustum as (Fig.~\ref{fig:3}): 


\begin{equation}\label{eq:mirror}
\begin{aligned}
    & P_{m} = K_m \, P \, \begin{bmatrix} \rf{O}{OR} & \pf{O}{OR} \\ \mat{0}^\top & 1  \end{bmatrix} \begin{bmatrix} R_x(\pi) & \begin{bmatrix} 0 \\ 0 \\ D \end{bmatrix} \\ \mat{0}^\top & 1  \end{bmatrix}, \\
    & K_m  = \begin{bmatrix} 1 \;\; & 0 \;\; & 0 \\ 0 \;\; & -1 \;\; & 0 \;\; \\ 0 \;\; & 0 \;\; & 1 \end{bmatrix} K_o .
\end{aligned}
\end{equation}

In Eq.~\ref{eq:mirror}, the optical center of the observer frustum is rotated by the amount $\pi$ around the $x$ axis, and translated by the amount $D$ along the principle ray of the frustum. Distance $D$ is approximated as the Euclidean distance between the camera center, and an arbitrary point on the surface of the robot that is acquired by colliding the gaze cursor with the spatial map of the AR scene. The distance is merely used as a reference to position the optical center of the reversed frustum, and does not affect the rendering content in the reflective-AR display. To compute the reversed frustum's intrinsic matrix $K_m$, the $y$-axis of the image plane is flipped according to Eq.~\ref{eq:mirror}. Lastly, to give rise to a mirror-like AR display, the 3D virtual structures are projected into the image plane with the imaging geometry $P_m$, thus, enabling joint visualization of real and virtual in the reflective display.

Since the reflective AR displays are constructed based on the imaging geometry of the observer frustum, we adopted the same axis convention used in the computer vision community. We set the z-axis in the direction pointing away from the camera, along the principal ray, connecting the origin to the principal point on the image plane.

\subsection{Augmented Reality Assistance for Robot Set Up} \label{sec:method:assistance}
After the registration transformation is established between real and virtual worlds, a collision-free and safe virtual robot configuration can be presented to the medical assistant. The desired configuration can either be estimated via the inverse kinematics of the robot, or can be adjusted interactively using the virtual robot and the patient position on the surgical bed. The robot set up is then performed in joint-by-joint steps, following the virtual planning.

\section{EXPERIMENTAL RESULTS}
\begin{figure*}
  \centering
  \includegraphics[width=\textwidth]{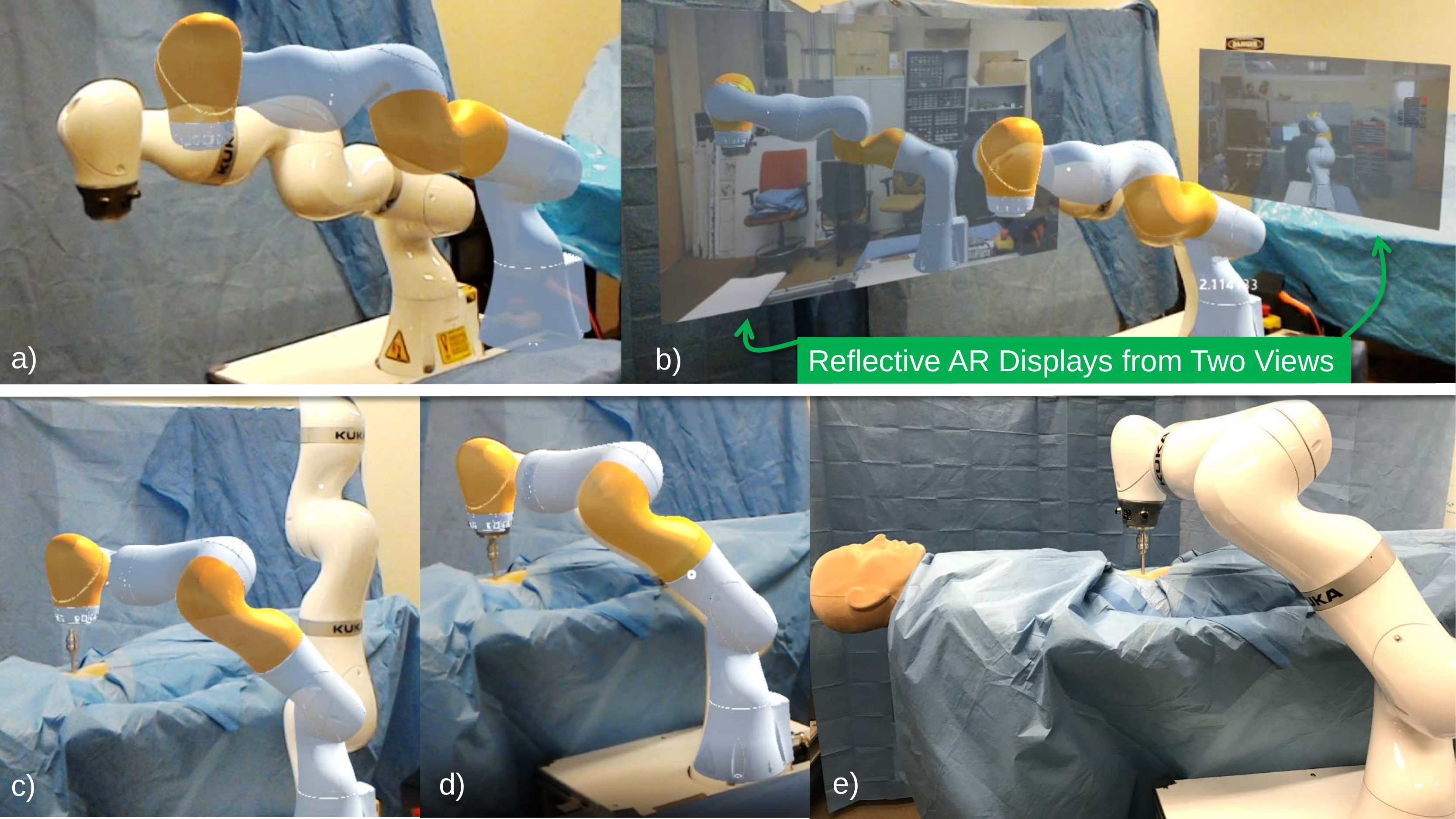}
  \caption{During the surgical AR experience, the virtual model of the robot is first visualized at a known configuration (\textbf{a}). The alignment between the real and virtual is established in multiple views via reflective-AR displays (\textbf{b}). Once the 6 DOF rigid-body transformation is identified between the real and virtual content, a virtual robot is rendered into the scene at a safe surgical configuration (\textbf{c}). The robot assistant can then align the robot with the virtual counterpart (\textbf{d}), and dock it to the trocar (\textbf{e}).}
    \label{fig:4}
\end{figure*}

We first evaluate the ViRAAl strategy using a virtual and real robotic manipulator. Next, in a simulated surgical setup, we assess the errors in moving the robot joints to achieve a desired joint configuration using AR guidance where a trocar must be inserted at a mannequin's umbilicus (Fig.~\ref{fig:4}). The umbilicus is commonly chosen as a robotic port and remote center of motion (RCM) for abdominal surgery. Training for port and trocar placement in umbilicus and optimal docking of the robot has a steep learning curve~\cite{chang2014patient, iranmanesh2010set}.

\subsection{System and Design}
For the experiments we used a 7 DOF KUKA LBR Intelligent Industrial Work Assistant (iiwa) 7 R800 redundant robot manipulator (KUKA AG, Augsburg, Germany). The joint configuration and end-effector pose of the robot was obtained through a ROS interface~\cite{safeea2019kuka}. It is important to note that our solution is designed to address challenges in surgical settings, and the KUKA arm was merely used as an exemplary robot that was available for this research. The AR environment was delivered by a first-generation Microsoft HoloLens OST-HMD (Microsoft, Redmond, WA).

\subsection{Alignment of Virtual-to-Real}\label{sec:subsec:results}
The ViRAAl strategy is evaluated by aligning the virtual and real robots with and without a reflective-AR display. Each experiment is repeated $10$ times for $4$ users. The error measurements are presented in Table~\ref{table:1}. 
We did not incorporate an external marker-based tracking approach as the base-line since marker tracking exhibits high errors due to propagation, and does not include the user in the loop, i.e. it only determines the registration error and not the augmentation error in the user's view. Instead, to quantify the amount of misalignment, for each iteration we located three pairs of distinct 3D landmarks on the surfaces of both real and virtual robots. These points were identified interactively by intersecting rays from the OST-HMD to the landmark using a gaze cursor.

We define each ray $i$ using the position of the user's head $\mat{h}_i$, and the unit direction vector $\mat{u}_i$ from the head to the annotated landmark on the spatial map of the environment. The landmark $\mat{x}_i^{*}$ is estimated in a least-squares fashion from the intersection of two rays as:
\begin{equation}
        \mat{x}_i^{*} = \underset{\mat{x}\in\mathbb{R}^3}\argmin{ \sum_{i=1}^2 \Vert (I_3 -\mat{u}_i \mat{u}_i^{\top}) (\mat{x} - \mat{h}_i)\Vert^2}.
\end{equation}

To quantify the error of our ground-truth measurement mechanism using 3D landmarks, we computed the Euclidean distance between different sets of targets on an optical table for a total of $12$ times. We selected four combinations of landmarks which were $5$\,cm, $10$\,cm, $15$\,cm, and $20$\,cm apart. The average error for measuring distances using AR annotations was $3.6$\,mm.

\begin{table*}
\centering
\caption{Mean and standard deviation of misalignment errors in mm.}
\label{table:1}
\begin{tabular}{l|c|c|c|c}
\rowcolor[HTML]{BBBEEA} 
\textbf{Alignment Method} & $(\overline{t_x}, \sigma_{t_x})$ & $(\overline{t_y}, \sigma_{t_y})$ & $(\overline{t_z}, \sigma_{t_z})$ & $(\norm{\overline{t}}_2, \norm{\sigma_t}_2)$ \\ \hline
\textbf{ViRAAl} & $(17.4, 16.1)$ & $(11.9, 6.24)$ & $(21.6, 16.5)$ & $(30.2, 23.9)$ \\
\textbf{ViRAAl + Reflective-AR Display} & $(9.00, 5.64)$ & $(10.3, 7.45)$ & $(9.18, 5.77)$ & $(16.5, 11.0)$
\end{tabular}
\end{table*}

\begin{figure*}[t]
  \centering
  \includegraphics[width=0.70\textwidth]{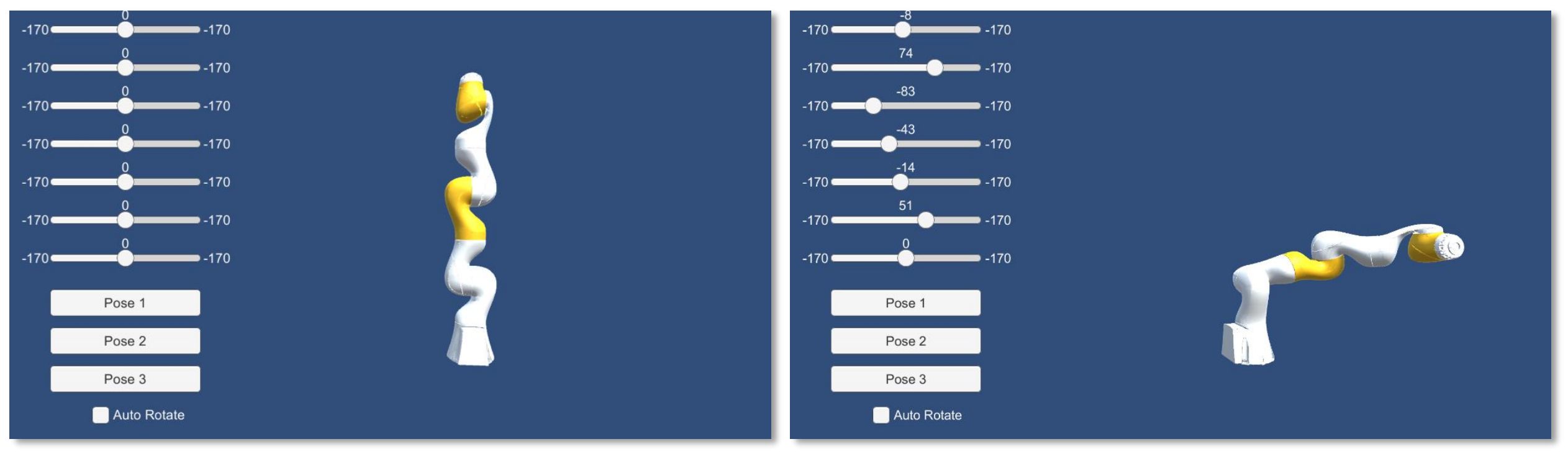}
  \caption{Interactive fixed display demonstrating two different target joint configurations}
    \label{fig:fixeddisplay}
\end{figure*}

For virtual-to-real object alignment, the results in Table~\ref{table:1} indicate a total error of $16.5 \pm 11.0$\,mm when using the reflective-AR display, and $30.2 \pm 23.9$\,mm when using AR without the additional mirror view. To demonstrate the change in alignment error when averaging multiple alignment transformations on the $SE(3)$ manifold as presented in Sec.~\ref{sec:method:ViRAAl}, we computed the average transformation given Eq.~\ref{eq:averagerotation} and Eq.~\ref{eq:averagetranslation} when using the AR reflective display. This experiment yielded a total error of $11.3 \pm 1.01$\,mm, which is lower than each individual alignment trial.

\begin{figure}
  \centering
  \includegraphics[width=\columnwidth]{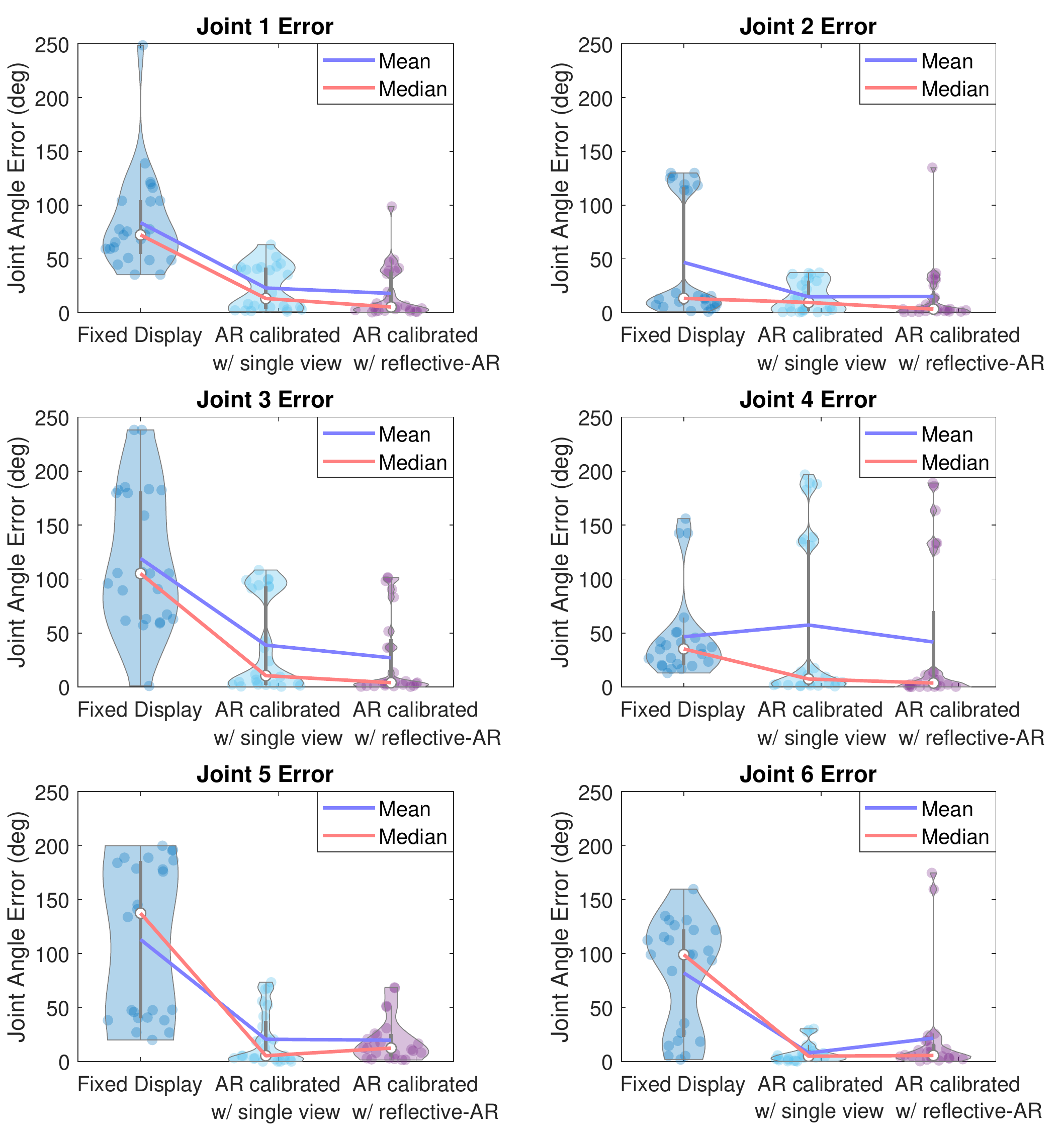}
  \caption{Distribution of errors evaluated for each joint separately when guided by \textit{1)} non-immersive fixed display, \textit{2)} AR calibrated without reflective-AR display, and \textit{3)} AR calibrated with reflective-AR display.}
    \label{fig:jointerror}
\end{figure}

\subsection{Augmented Reality for Robot Set Up: Accuracy Analysis}\label{sec:subsec:ARRobAccuracy}

\begin{figure}[t]
  \centering
  \includegraphics[width=\columnwidth]{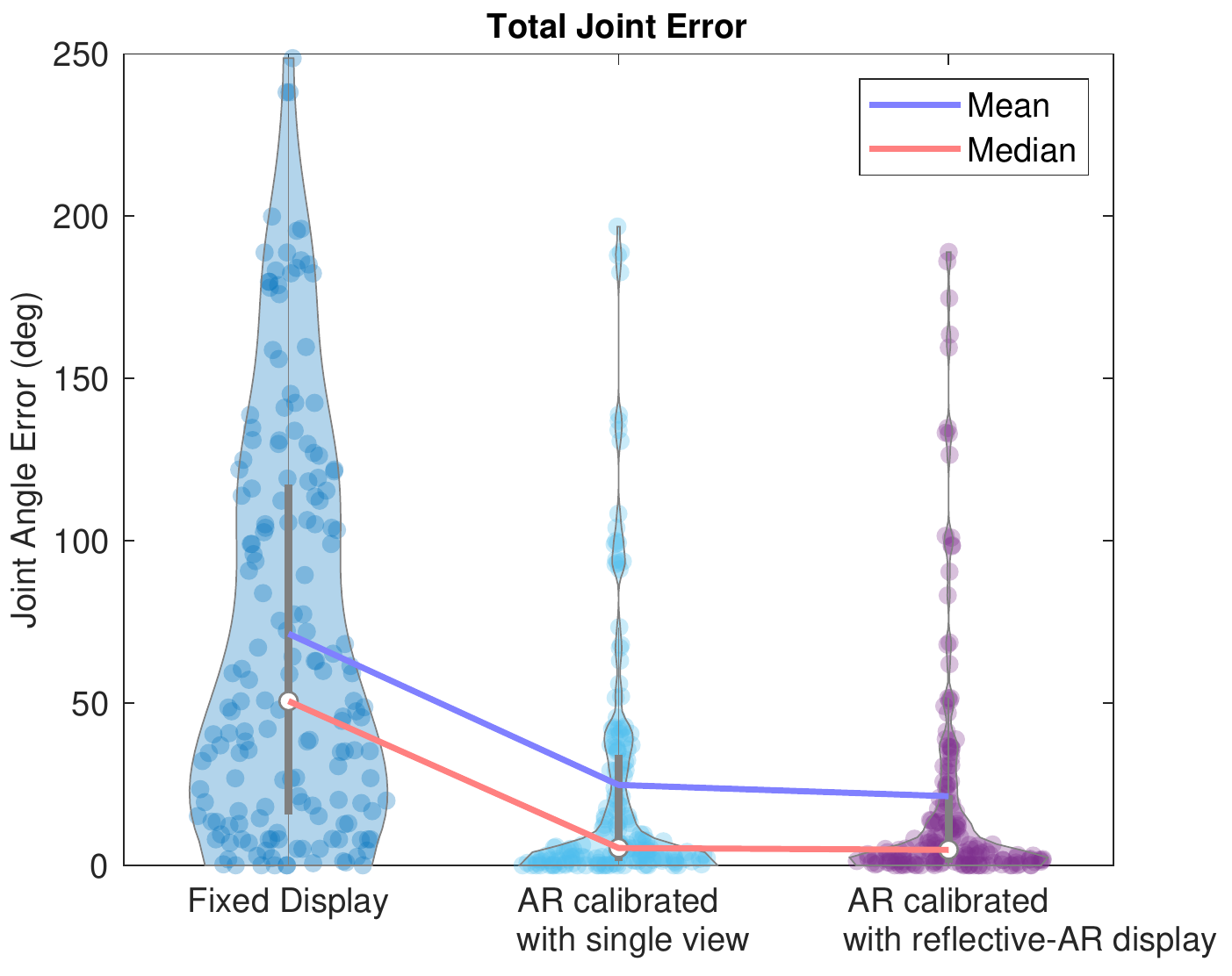}
  \caption{Total error distribution for all joints using guidance by \textit{1)} non-immersive fixed display, \textit{2)} AR calibrated without reflective-AR display, and \textit{3)} AR calibrated with reflective-AR display.}
    \label{fig:totalerror}
\end{figure}

During a simulated robot-assisted trocar placement, $U=8$ users moved the robot joints to achieve different target joint configurations. Each user performed this task with the guidance from \textit{1)} AR, \textit{2)} AR with reflective display, and \textit{3)} interactive fixed display, all in randomized orders. The fixed external monitor was used as a non-immersive baseline, displaying the desired robot configurations to the user at all time during the execution, and allowing the users to interact with the visualization by rotating or scaling (Fig.~\ref{fig:fixeddisplay}). The users performed each test three times, and each time with a different target joint. To set up AR we used the average Euclidean transformation over $N=3$ trials as described in Eq.~\ref{eq:averagerotation} and Eq.~\ref{eq:averagetranslation}.  Errors in joint angles are demonstrated in Fig.~\ref{fig:jointerror}. The total errors are shown in Fig.~\ref{fig:totalerror}. The violin plots show the distribution of error within each guidance method.

\begin{figure}
  \centering
  \includegraphics[width=\columnwidth]{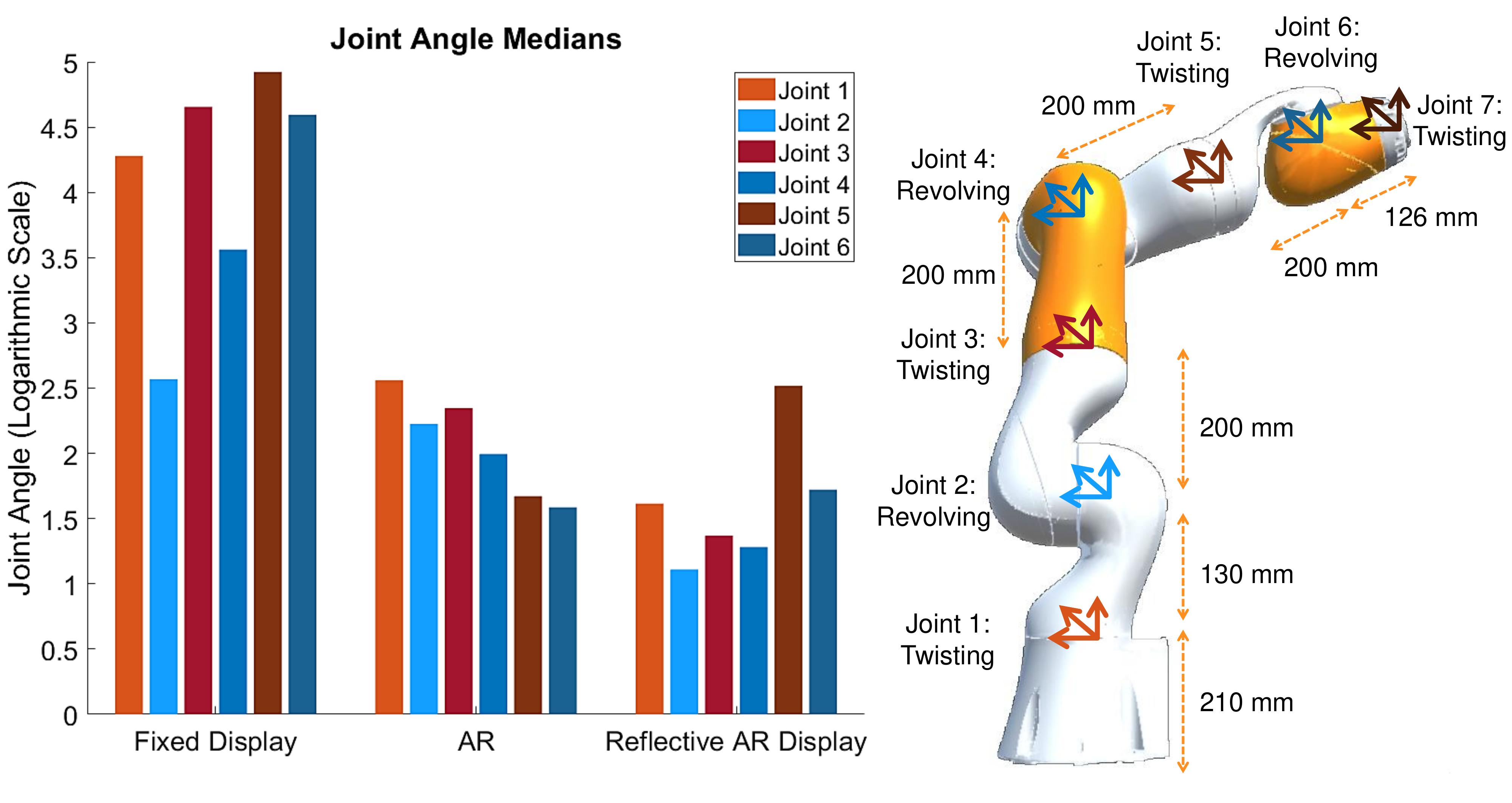}
  \caption{The plot demonstrates an abstract comparison between the errors contributed by each joint. Since AR-based approaches yielded substantially smaller errors, we used Logarithmic scale for optimal visualization and comparison of errors with different orders of magnitude. Revolving joints with even indexes are shown in blue, and twisting joints with odd indexes are shown in red.}
    \label{fig:jointrank}
\end{figure}

The statistics for the overall error in joint positioning using each guidance techniques is compared in Table~\ref{table:Accuracycomparison}. Statistical significance measures are shown in Table~\ref{table:P-value}. In Fig.~\ref{fig:jointrank} we present a comparison between the errors from each joint, and highlight the errors for twisting and revolving joints. A joint is characterized as twisting if the axis of rotation is parallel to the robot link, and revolving if the axis of rotation is orthogonal to the robot link.

\begin{table}
\caption{Comparison of the error for re-positioning the robot joints in degree units}
\label{table:Accuracycomparison}
\begin{tabular}{l|c|c|c|c|c}
\rowcolor[HTML]{BBBEEA} 
\textbf{Joint Error} & Mean & Median & Min & Max & \multicolumn{1}{l}{\cellcolor[HTML]{BBBEEA}Std} \\ \hline
\textbf{Reflective AR Display} & $23.7$ & $4.93$ & $0.02$ & $188$ & $41.1$ \\
\textbf{AR} & $26.8$ & $5.88$ & $0.00$ & $197$ & $42.4$ \\
\textbf{Fixed Display} & $71.4$ & $50.6$ & $0.64$ & $249$ & $61.8$
\end{tabular}
\end{table}

\begin{table*}
\centering
\caption{P-values for each individual joint, as well as for all joints combined.}
\label{table:P-value}
\begin{tabular}{l|c|c|c|c|c|c|c}
\rowcolor[HTML]{BBBEEA} 
\textbf{P-value} & \textbf{Joint 1} & \textbf{Joint 2} & \textbf{Joint 3} & \textbf{Joint 4} & \multicolumn{1}{l|}{\cellcolor[HTML]{BBBEEA}\textbf{Joint 5}} & \textbf{Joint 6} & \textbf{Total} \\ \hline
\textbf{Reflective AR Display / AR} & $0.61$ & $0.50$ & $0.23$ & $0.71$ & $0.59$ & $0.27$ & $0.54$ \\
\textbf{Reflective AR Display / Fixed Display} & $0.12\mathrm{e}{-6}$ & $0.15\mathrm{e}{-1}$ & $0.24\mathrm{e}{-6}$ & $0.77$ & $0.22\mathrm{e}{-6}$ & $0.76\mathrm{e}{-4}$ & $0.12\mathrm{e}{-15}$ \\
\textbf{AR / Fixed Display} & $0.16\mathrm{e}{-6}$ & $0.39\mathrm{e}{-1}$ & $0.62\mathrm{e}{-4}$ & $0.87$ & $0.12\mathrm{e}{-6}$ & $0.58\mathrm{e}{-7}$ & $0.20\mathrm{e}{-17}$
\end{tabular}
\end{table*}


\begin{table*}
\caption{Time required for ViRAAl and re-positioning the robot joints in {minute:second} units}
\label{table:ExecutionTimeComparison}
\centering
\begin{tabular}{l|c|c|c|c|c|c|c|c|c|c|c}
\cline{2-6} \cline{8-12}
 & \multicolumn{5}{c|}{\cellcolor[HTML]{BBBEEA}\textbf{Execution Time}} & \multicolumn{1}{l|}{} & \multicolumn{5}{c|}{\cellcolor[HTML]{BBBEEA}\textbf{Alignment Time}} \\ \cline{2-6} \cline{8-12} 
\rowcolor[HTML]{BBBEEA} 
\cellcolor[HTML]{FFFFFF}\textbf{} & Mean & Median & Min & Max & Std & \cellcolor[HTML]{FFFFFF} & Mean & Median & Min & Max & Std \\ \cline{1-6} \cline{8-12} 
\textbf{Reflective AR Display} & $2:00$ & $1:50$ & $0:39$ & $3:47$ & $0:50$ &  & $4:32$ & $4:42$ & $2:06$ & $6:59$ & $1:41$ \\
\textbf{AR} & $1:31$ & $1:23$ & $0:22$ & $5:06$ & $0:59$ &  & $2:29$ & $1:52$ & $1:11$ & $4:26$ & $1:12$ \\
\textbf{Fixed Display} & $1:34$ & $1:35$ & $0:27$ & $3:20$ & $0:48$ &  & - & - & - & - & -
\end{tabular}
\end{table*}

\subsection{Augmented Reality for Robot Set Up: Time Analysis}\label{sec:subsec:ARRobTime}

Table~\ref{table:ExecutionTimeComparison} presents the observed time for all eight users. Guidance using AR and AR reflector both required registration between the virtual and real content. Therefore, each time the users were given a unique joint configuration target, prior to moving the real robot, the user aligned the virtual model of the robot with its real counterpart three times $\left\{\tf{R}{V} (i)\right\}_{i=1}^{3}$. Using the ViRAAl approach presented in Sec.~\ref{sec:method:ViRAAl}, the transformation $\overline{\tf{R}{V}}$ which expressed the geometric mean in $SE(3)$ was computed. The average transformation was applied to the AR scene to register the origins of the real and virtual environments, and enable AR guidance. Table~\ref{table:ExecutionTimeComparison} also presents the time required for ViRAAl with and without AR reflectors for all users.


\section{DISCUSSION}
The experimental results indicated an improved alignment when using a reflective-AR display. The L2-norm average misalignment error in this case was $16.5 \pm 11.0$\,mm, and showed improvement compared  to $30.2 \pm 23.9$\,mm error when no reflective displays were used. Averaging the transformations on $SE(3)$ manifold yielded an even lower error of $11.3 \pm 1.01$\,mm. This alignment error does not seem sufficient for tasks that require high accuracy such as defining biopsy targets in AR, however seems acceptable for providing intuition during robotic arm set up. Using more than one reflective-AR display did not improve the alignment due to two main reasons. First, the limited field of view of Microsoft Hololens prohibited optimal view of multiple mirrors in their frustums simultaneously when standing in close proximity to the robot. Second, the poor quality of SLAM-based tracking and the unreliable spatial map of the HMD resulted in drifts, hence achieving alignment consensus in all views became challenging. Larger field of view, reliable head tracking, enhanced form factor, enhanced gesture input, and eye tracking capabilities can greatly improve the current limitations.

We evaluated the results in Table~\ref{table:1} based on a novel user-in-the-loop concept. Alternatively, these errors can be evaluated by using an external navigation system and fiducials. However, the latter will exclude the user and will only evaluate the registration error, instead of the augmentation error. Considering the large improvement $(>45\%)$ reported across multiple trials in Table~\ref{table:1}, we expect the conclusion will not change when using optical navigation. 

For modern surgical platforms, the first assistants are trained to set up, dock, tear down, and re-configure the robot using extensive pre-operative E-learning or instructor-led training. In addition to the training that they receive by the manufacturer, general guidelines, and demonstrations in the form of text or visualizations are available in the operating room. In our experiments, to exclude the bias of training, we substituted the conventional training-based approach by recruiting users with no background in setting up surgical robots and focused on demonstrating the effectiveness of intra-operative AR guidance. Comparison with base-line training is a subject of future work, which requires randomized studies in clinical settings with larger populations.

The alignment between the real and virtual, which establishes the registration, is an entirely user-dependent step as the registration chain implicitly takes into account the internal relations between the user’s eyes, AR camera on the headset, and the AR displays. Since human visual systems are different, these relations differ, and consequently cannot be done in a user-out-the-loop setting. Therefore, automatic approaches will not generalize.

Fig.~\ref{fig:totalerror} presented the misalignment errors for bringing the real robot to a desired configuration. The expected accuracy for the robotic set up depends on the design, number of arms, and number of joints of the particular surgical system. These parameters all vary depending on the surgical use case. The higher the number of joints and arms, the higher the chance for collision; therefore, higher accuracy is demanded for a more complex system. This step during the AR-assisted workflow, which involves the alignment of a real object to virtual, can in future leverage from multi-view strategies by using collaborative AR devices or external cameras.

Several AR publications have shown that time-saving of AR cannot be quantified immediately with dedicated user studies~\cite{dunser2011evaluating}, partly because of the unfamiliar interface and exposure to additional information. Time-saving only manifests after the user is proficient with the system. We hypothesize that while there is overhead in setting up AR, the rate of failure/collision would drop leading to a net reduction of overall junk-time. 
The Riemannian averaging in Eq.~\ref{eq:averagerotation} and Eq.~\ref{eq:averagetranslation} can increase this set up time, but is a one-time process which takes place before intervention and can result in a more accurate fusion of information and improved AR experience.
We hope AR assistance minimizes the training time and allows operators to verify and inspect the proper alignment of robot arms, both quantitatively and visually.

HMD-based AR is challenged by recurrent estimation of transformations and corrections such as user-to-display and drift, respectively. There is a great wealth of literature directed towards addressing these issues~\cite{tuceryan2002single}, which our solution benefits from as new hardware becomes available. It should be noted that drift will be quite limited because the working volume near the patient bed and the robot is restricted.

Our proposed approach enables the co-registration between the real and virtual spaces and delivers spatially-aware AR. We also demonstrated the application of ViRAAl for AR guidance during minimally-invasive robotic surgery. 
The estimation of the overall registration greatly benefits from the averaging strategy presented in  Eqs.~\ref{eq:averageRotationdefinition}-\ref{eq:averagetranslation}, which are suggested to compute the mean transformation that satisfies the properties of $SE(3)$ manifold. This average estimate is a rigid transformation that has the shortest distance to all other estimates around the true pose.

The range of errors exhibited by all intra-operative guidance methods, particularly by the non-immersive fixed display, prove the complexity and importance of this problem for robot manipulation. We computed p-values and compared all pairs of methods in Table~\ref{table:P-value} to identify the most effective assistive approach. Statistical significance was considered if $p < 0.05$. Results suggested that guidance using AR with and without reflective display yielded significantly lower errors compared to non-immersive fixed display. The AR guidance approach using reflective displays outperformed the AR system with no mirrors, however in this comparison statistical significance was not achieved.

In Fig.~\ref{fig:jointrank}, we compared the error contributed by revolving and twisting joints separately. Results indicated that re-positioning of revolving joints in all three guidance methods are consistently more accurate than the twisting joints. We hypothesize that the higher error is the result of the inherent symmetry in twisting motion that may lead to ambiguities.

\section{CONCLUSION}
In this manuscript we presented a novel multi-view strategy to align virtual and real content, and demonstrated an application of it for improving surgical robotic workflows. The reflective-AR displays were introduced to eliminate the 3D scale ambiguities and improve the AR scene realism. We have demonstrated an AR interface that accommodates multiple reflective displays, and allows the users to scale the images within their viewing frustum~\cite{fotouhi2019interactive}.

The virtual-to-real registration approach, ViRAAl, is an interactive and user-specific method that calibrates the real and virtual worlds directly to the user's display. No external camera or tracking system, other than the HMD itself, is used in order to keep maximum flexibility and transferability of the system into different surgical environments.

Seamless overlays of virtual content onto the reflective AR displays are achieved by placing virtual cameras at the optical centers where the images were acquired, hence allowing to render virtual and real from an identical imaging geometry. The reflective displays require a static scene, therefore, are suited for aligning virtual-to-real, and not vice-versa. 

The focus of this work is beyond the KUKA robot and its redundant design; it is instead on complex surgical platforms with multiple arms and various joints. The task of alignment is expected to be more difficult for redundant manipulators with more joints. Nonetheless, we expect AR to provide an effective guidance mechanism to reconfigure complex redundant arms at the bedside. 

Surgical robots are only certified to be controlled by the surgeon in its active mode, and due to safety reasons, their set up by the surgical staff are performed entirely manually. Our solution is designed around this concept of full manual interaction. It should be noted that our contribution is not on computing joint configurations that minimize collision, but instead we show that if such configuration exists, then with the support of AR it can be manually achieved by the first assistant.

A user interface such as the AR reflectors, can accelerate interaction during surgery. By measuring the exact time of the staff this could be validated in future work. We believe that the proposed alignment strategy can extend to other realms of computer-assisted surgery, namely for surgical training and AR guidance during image-guided therapies.

\addtolength{\textheight}{-12cm}  


\bibliographystyle{IEEEtran}
\bibliography{main}

\end{document}